\RequirePackage[hyphens]{url}

\documentclass[a4paper,twoside]{article}

\usepackage[utf8]{inputenc}
\usepackage[hidelinks]{hyperref}
\usepackage{epsfig}
\usepackage{subcaption}
\usepackage{calc}
\usepackage{amssymb}
\usepackage{amstext}
\usepackage{amsmath}
\usepackage{amsthm}
\usepackage{multicol}
\usepackage{pslatex}
\usepackage{apalike}
\usepackage{mathabx}
\usepackage{SCITEPRESS}     

\begin{document}

\title{The Person Index Challenge:\\Extraction of Persons from Messy, Short Texts}

\author{\authorname{Markus Schröder\sup{1,2}, Christian Jilek\sup{1,2}, Michael Schulze\sup{1,2} and Andreas Dengel\sup{1,2}}
\affiliation{\sup{1}Smart Data \& Knowledge Services Dept., DFKI GmbH, Kaiserslautern, Germany}
\affiliation{\sup{2}Computer Science Dept., TU Kaiserslautern, Germany}
\email{\{markus.schroeder, christian.jilek, michael.schulze, andreas.dengel\}@dfki.de}
}

\keywords{
Person Index,
Extraction,
Short Text
}

\abstract{
When persons are mentioned in texts with their first name, last name and/or middle names, there can be a high variation which of their names are used, how their names are ordered and if their names are abbreviated.
If multiple persons are mentioned consecutively in very different ways, especially short texts can be perceived as ``messy''.
Once ambiguous names occur, associations to persons may not be inferred correctly.
Despite these eventualities, in this paper we ask how well an unsupervised algorithm can build a person index from short texts.
We define a person index as a structured table that distinctly catalogs individuals by their names.
First, we give a formal definition of the problem and describe a procedure to generate ground truth data for future evaluations.
To give a first solution to this challenge, a baseline approach is implemented.
By using our proposed evaluation strategy, we test the performance of the baseline and suggest further improvements.
For future research the source code is publicly available.
}

\onecolumn \maketitle \normalsize \setcounter{footnote}{0} \vfill

\section{\uppercase{Introduction}}
\label{sec:introduction}

\noindent
In the Western world, it is common that persons have a first name (forename, given name), last name (surname, family name) and optionally additional names like middle names.
When individuals are mentioned in texts, there can be a high variation which of their names are used, how their names are ordered and if their names are abbreviated.
For example, ``John Fitzgerald Kennedy'',  ``John'',  ``Kennedy, J F'' and  ``J. Kennedy'' are variations that refer to the same person.

Once people share equal names, references can easily become ambiguous even in smaller groups.
In these cases, readers try to disambiguate them with additional context information given in texts.
However, especially short texts (or text snippets) often lack a regular grammar, have only few statistical signals and are rather ambiguous \cite{Hua2015}.
Thus, in a worst-case scenario, correct associations to individuals are impossible to infer.
Nevertheless, eligible persons could be suggested.

Often, we encounter unstructured (short) texts where several persons are mentioned but we do not have an index which lists them clearly. 
We define such a person index as a structured table that distinctly catalogs individuals by their names.
However, arranging this index becomes a challenging task for humans and especially for machines, when we consider the previously mentioned eventualities:
optional (middle) names, name variations, ambiguities and short texts.
In particular, the messiness of texts makes this challenge more difficult.
We classify such texts as ``messy'' if mentioned person names do not follow a particular pattern (or structure), i.e. the data quality is rather low.
Still, an initial suggestion for an index could be calculated by an unsupervised algorithm to reduce the manual effort considerably.
Because such a method's performance is still unclear in our described scenario, we ask the following research question:
\textit{How well can an unsupervised algorithm build a person index from a set of messy, short texts?}

In order to answer this question, we will design a procedure that generates ground truth data to conduct evaluations with proposed solutions.
The generator is able to produce a list of short texts referring to persons in various forms and generates an index of persons that has to be discovered.
To acquire first performance results, we propose a baseline algorithm. 
This work does not intend to provide a novel procedure to solve the person index challenge in the best way possible.
Instead, the contributions of this paper are the following:
\begin{itemize}
	\item a formal definition of the problem,
	\item a procedure that generates ground truth data for this specific challenge,
	\item an evaluation strategy to assess the quality of algorithms that try to solve it
\end{itemize}
For future research the source code of the generator, the baseline algorithm and the evaluator is publicly available at GitHub\footnote{\url{https://github.com/mschroeder-github/person-index}}. 
This paper is structured as follows:
The next section will formally define the problem of building a person index from texts.
After the discussion of related work (Section \ref{sec:relwork}), we suggest a data generator that produces ground truth for future evaluations (Section \ref{sec:gt}).
Section \ref{sec:approach} describes a baseline approach for the given problem.
Our evaluation strategy in Section \ref{sec:eval} shows first performance results.
Section \ref{sec:concl} concludes the paper and describes future work.

\subsection{Problem Definition}
\label{sec:prob}

The problem is a specific form of named entity recognition (NER) and disambiguation (NED).
At first, persons with their names have to be recognized in text snippets as usual.
However, it is important to decide which name is first name, last name and middle name to fill the person index correctly.
Although we are aware that there are persons having more than one middle name, like for example ``J. R. R. Tolkien'', we simplified our problem to correspond more to familiar industrial scenarios. This is also the reason why we currently focus on Western names only.

The disambiguation of persons can not be done with a preexisting person knowledge base in our scenario since there is no such source in advance.
Instead, disambiguation has to be done by examining collected entries in the person index.
Because of a particular uncertainty in this process, there can be ambiguous person references.
What follows is a formal specification of the described problem:

A person is a tuple $p_j := (fn, mn, ln)$ containing a first name ($fn$), a last name ($ln$) and an optional middle name ($mn$).
Given a set of short texts $t_i \in T$, the challenge is to extract all distinct persons $p_j \in P$ from 
their texts mentions such that (as far as possible) their names are at full length.
In order to know which person was mentioned in which text, a relation $(t_i, p_j) \in R \subseteq T \times P$ has to be provided.
If references are ambiguous and there is no way to disambiguate them, the relation $(t_i, r, P_A) \in A$ with $P_A \subseteq P$ shall capture that in a short text $t_i$ -- due to a substring $r$ (reason) -- a set of persons $P_A$ are possibly mentioned.
We distinguish between $P$ and $A$ to ease the later evaluation of correctly found unambiguous persons versus ambiguous ones.
\\
\\
\noindent
\textbf{Example.} Given the texts $t_1=$``Baker$\dlsh$Thompson LS-Z-U'', $t_2=$``mail to Chief Morgan (Wilson), [remove Baker, Robert]'' and $t_3=$``Wilson, M.; Susan Lea Baker'', the following persons can be discovered: 
$p_1=(Robert,\emptyset,Baker)$, $p_2=(Wilson,\emptyset,Morgan)$, $p_3=(\emptyset,\emptyset,Thompson)$ and $p_4=(Susan,Lea,Baker)$.
This leads to the relation $ R = \{(t_1, p_3), (t_2,p_2), (t_2, p_1), (t_3, p_2), (t_3, p_4) \}$.
Because ``Baker'' is ambiguous in $t_1$, we state $ A = \{ (t_1, \textrm{Baker}, \{ p_1, p_4 \}) \}$.

\section{\uppercase{Related Work}}
\label{sec:relwork}

\noindent In the area of natural language processing (NLP), information extraction (IE) \cite{DBLP:journals/semweb/Martinez-Rodriguez20} and its subfield named-entity recognition (NER) \cite{ner} are well known disciplines.
Usually, NER models are trained to recognize certain entity types in texts such as locations, organizations or persons.
They can be divided into two categories whether they know all possible entities upfront, such as in \cite{DBLP:conf/ijcai/SongWWLC11}, or they have to detect them blindly.
Our scenario belongs to the second category with focus on the recognition of person entities in texts.
Instead of commonly written documents like books or websites, we further limit the corpus to be only short texts \cite{Hua2015}.

NER for SMS \cite{ek2011named} can also be considered as short texts since they do not necessarily follow a regular grammar.
The authors' supervised algorithm is pre-trained with an annotated SMS corpus and additionally supported with gazetteer lists.

There are many similar works which try to recognize entities in short texts.
However, they usually do not form an index of canonical forms as stated in the problem definition.
More similar to our scenario is named-entity normalization (NEN).
Besides usual recognition, it integrates a normalization process to assign unique identifiers to entities.

A person normalization problem is solved in \cite{DBLP:conf/sigir/JijkounKMR08} by using within-document reference resolution in user generated contents.
If the person cannot be disambiguated with a Wikipedia lookup -- which is a typical case in our scenario -- the person's surface form is used instead.
In their appendix, a person name matching algorithm is described which uses heuristics and fuzzy matching.
Another normalization strategy in tweets \cite{DBLP:conf/acl/LiuZZFW12} finds overlapping tokens in the entities' names. 
The canonical form of an entity is the one with the maximum words.

Instead of persons, gene mentions were recognized and normalized \cite{DBLP:conf/ismb/Cohen05}.
For normalization, a dictionary is generated that contains many orthographic variants how genes could be mentioned.
In a similar way, sCooL \cite{DBLP:conf/cts/JacobJZM14} normalizes academic institution names. 
Normalization is also utilized to improve question answering \cite{DBLP:conf/ecir/KhalidJR08}.

To the best of our knowledge, there is no work that proposed the problem of building a person index (or a similar index) as we did.

\section{\uppercase{Ground Truth Generator}}
\label{sec:gt}

In order to be able to evaluate our and future approaches, we propose a generator that produces ground truth data.
Considering the problem definition in Section \ref{sec:prob}, the generator produces four comma-separated values (CSV) files: 
a set of short texts mentioning persons in various ways ($T$), a person index that lists all mentioned persons ($P$), a list of relations that relate short texts to persons ($R$) and a disambiguation list ($A$).
As input, our procedure expects a catalog of first names and last names.
Additionally, several parameter settings can be passed to customize the generation behavior such as a random seed to control randomness and quantities to adjust how many persons and short texts should be generated.

In order to control the degree of ambiguity, the user can decide how many groups of persons will share either first name or last name.
Also, the size of these groups can be specified.
For example, if the degree of ambiguity is set to two, two groups of two people each share a last name while other two groups of two persons each share a first name.
Our generator starts with the creation of persons having ambiguous names.
To avoid producing more ambiguity later, the selected ambiguous first names and last names are not used again.
As an example, Robert Baker and Susan Lea Baker were generated in the ambiguous last name group ``Baker''.
The rest of the persons are generated straight forward without using any name twice.
Persons with middle names are generated as well.
Their number can be adjusted in the generator's settings.
In these cases, another first name is randomly picked as a middle name.
This way, ``Susan Lea Baker'' was produced in our example.

After the person index is completed, the short texts are generated.
The generation procedure is heavily inspired by concrete data observed in an industrial scenario where spreadsheets were completed by individuals over years.
Since copy\&paste was often used to transfer names from several information systems and files, various name variations can be found in the data.  
That is why each generated short text (representing a spreadsheet cell) will mention a single person or a group of people at random.
To ensure that the data is messy in a similar way to the observed data, every person is mentioned using a different variation.
\autoref{table:patterns} lists fourteen patterns the generator uses to refer to a person.
In the patterns the variables for first name ($fn$), middle name ($mn$) and last name ($ln$) are used.
The function $letter(name)$ returns the first letter of a name while $lc(name)$ converts a name to lower case.
Following procedures generate random strings:
$department()$ returns a string that looks like a description of a department while $rnd(n)$ generates a random $n$-length alphabetic string.
$note()$ randomly selects a short note from a list like ``old'', ``TODO'' and ``remember''.
The same way, $role()$ randomly picks a role description from a list like ``Executive'', ``CEO'' and ``Chief''.
The `$\dlsh$' symbol indicates a new line.
To demonstrate how they will look like, the full name ``John Fitzgerald Kennedy'' is used as an example.
Note that not all patterns mention all names completely which is captured with the FN (first name mentioned), MN (middle name mentioned) and LN (last name mentioned) columns.
This information is vital to foresee ambiguity later.
Moreover, some patterns contain additional information like department names, mails or notes to make texts more realistic and to potentially distract detection algorithms.
If the person has a middle name, the generator makes sure to use patterns 11 to 14.
The 11th pattern is always used first to ensure that the person's middle name was mentioned at least once.
\begin{table*}[t]
	\centering
	\caption{
		Patterns to generate mentions of persons in various ways which are demonstrated by an example.
		An `x' in FN, MN and LN indicates that first name, middle name or last name are fully mentioned in a pattern.
	}
\begin{tabular}{|c|l|l|c|c|c|}
		\hline
		Nr. & Pattern & Example & FN & MN & LN \\
		\hline
		\hline
		1 & $fn$ & John  & x &  &  \\
		2 & $ln$ & Kennedy &  &  & x \\
		3 & $fn$ $ln$ & John Kennedy  & x &  & x \\
		4 & $ln$ $fn$ & Kennedy John  & x &  & x \\
		5 & $ln$, $fn$ & Kennedy, John  & x &  & x \\
		6 & $ln$, $letter(fn)$. & Kennedy, J.  &  &  & x \\
		7 & $ln$ $department()$ & Kennedy US-Z-G  &  &  & x \\
		8 & $department()$$\dlsh$$ln$ $fn$ & US-Z-G$\dlsh$Kennedy John  & x &  & x \\
		9 & $ln$ $fn$ $<$$lc(ln)$@$rnd(5)$.$rnd(2)$$>$& Kennedy John $<$kennedy@xraok.nc$>$  & x &  & x \\
		10 & $note()$ $role()$ $ln$ $fn$ & new Admin Kennedy John & x &  & x  \\
		11 & $fn$ $mn$ $ln$ & John Fitzgerald Kennedy & x & x & x  \\
		12 & $fn$ $letter(mn)$. $ln$ & John F. Kennedy & x &  & x  \\
		13 & $letter(fn)$. $letter(mn)$. $ln$ & J. F. Kennedy &  &  & x  \\
		14 & $ln$, $letter(fn)$. $letter(mn)$.  & Kennedy, J. F. &  &  & x  \\
		\hline
		\hline
		Sum & - & - & 9 & 1 & 13 \\
		\hline
	\end{tabular}
	\label{table:patterns}
\end{table*}

If multiple persons are mentioned in a short text, they are separated by a random delimiter.
Additionally, their names can be surrounded by all kinds of brackets and quotes.
By this, we avoid that algorithms can simply split texts in a trivial way.
To give a short example, some short texts produced by our generator are listed below (separated by empty lines).
\begin{small}
	\begin{verbatim}
	[Sullivan, Arthur <sullivan@wnpql.to>];
	HR-X-C-N-G Brooks Alonso
	
	[Watson, L.];Campbell, Mikaela;Cooper VG-Z
	Isabella Adams - [Lee Zoey]
	
	{Chloe}; Martin, M.; Raquel Amanda Garcia;
	Myers Elijah
	
	Alice
	
	(RK-H Lewis Martin); new Executive Jones
	Olivia; Mariana Robinson; {Evans Mason}
	\end{verbatim}
\end{small}

During the text generation, our procedure records which person was unambiguously mentioned in a certain text in order to capture the relation $R$.
If, in contrast, a person has an ambiguous name, the relation $A$ is filled instead.

\section{\uppercase{Baseline Approach}}
\label{sec:approach}

\noindent
In this section, we describe our first approach to solve the person index challenge.
This work does not intend to provide a novel procedure to solve the person index challenge in the best way possible.
It is meant to present initial results and act as a comparative measure for future approaches.

The data's messiness is handled by our approach with some assumptions.
A guess about name capitalization and their ordering allows us to initially populate the person index.
Additionally, a first name gazetteer lookup is performed to potentially correct the name order.
For disambiguation there is no special handling.

First, our proposed procedure discovers persons in texts.
Commonly, named entity recognition (NER) is applied to identify entities such as organizations, locations and also persons in texts.
We therefore use OpenNLP\footnote{http://opennlp.apache.org/} which is a library that provides various natural language processing (NLP) algorithms.
Its module Name Finder\footnote{http://opennlp.apache.org/docs/1.9.2/manual/opennlp .html\#tools.namefind} allows to detect text entities of various types.
To do so, it requires a model which is pre-trained on a corpus in a specific language for a certain entity type.
In our algorithm, we utilize the person name finder model \verb|en-ner-person.bin|\footnote{http://opennlp.sourceforge.net/models-1.5/}, which was trained on an English corpus to detect names.
Although, this model was trained in a supervised manner, our baseline algorithm is still unsupervised since no ground truth (see previous section) was used to train it.  
Given a list of tokens, the NER model predicts with a certain probability if a sequence of tokens refers to an individual.
Unfortunately, the model can not decide which of the tokens are first name, middle name or last name.
That is why we make for now the following three provisional assumptions: 
if one token occurs, it is assumed to be a last name; 
if two tokens are found, the first token is last name and second one is first name; 
if three tokens are discovered, the token in the middle is presumed to be the middle name.
Doing this for given short texts yields to a list of persons where some of them occur several times.

Because of errors made in the detection, some persons get improper names such as symbols or letters.
We define a proper name as a name that starts with an upper case letter and ends with lower case letters.
To filter persons, we match their names with a regular expression.
In this process, duplicates are also removed.
Still, it is unclear, if first name, middle name and last name were correctly assigned.
Our solution is the usage of a first name gazetteer list.
We swap names accordingly if falsely assumed last names turn out to be first names.
However, this feature requires a previously compiled list.

Next, the relations ($R$) which relate short texts to persons are discovered.
We iterate again over all short texts and match tokens with (first and last) names of all persons in our index.
In case a single person was found, we state a relationship between the text and the person.
If multiple persons are matched, we make an entry in the ambiguity list ($A$).

Our approach has three features that can be activated independently.
First, OpenNLP's Name Finder allows to clear adaptive data\footnote{https://opennlp.apache.org/docs/1.9.2/apidocs/opennlp-tools/opennlp/tools/namefind/NameFinderME.html} that is collected during the text processing.
This can be done each time a new short text is processed which may improve the detection result.
Second, the probability measures of Name Finder's model can be used to filter uncertain detections.
If the probability is below $0.5$, tokens will not be regarded as names of a person.
Third, as already described, a first name gazetteer list can be used to swap names accordingly.

\section{\uppercase{Evaluation}}
\label{sec:eval}

\noindent
In this section, we present our evaluation strategy that assesses algorithms that try to solve the proposed person index challenge.
The ground truth $GT := (T, P, R, A)$ consists of short texts $T$, a person index $P$, a relation $R$ and an ambiguity list $A$.
Potential algorithms consume short texts in $T$ and output a person index $P_a$, a relation $R_a$ and an ambiguity list $A_a$.

First, we are interested in the algorithm's performance of building the person index. 
Therefore, the ground truth's index $P$ is compared with the algorithm's index $P_a$.
If all names of two given persons are identical, a correct match is assumed. 
Formally, an intersection of both sets can be calculated to get the matches $P_m := P \cap P_a$.
By this, we can calculate the algorithms precision and recall for assembling the person index: 
$$prec_P := \frac{|P_m|}{|P_a|}  \qquad \qquad recall_P := \frac{|P_m|}{|P|} $$

Second, we examine how often a correct mapping between short text and person are suggested.
This makes only sense for persons which are correctly found by the algorithm, namely $P_m$.
Thus, for each person $p_k \in P_m$, the relations\\$ \hat{R} := \{ (t_i, p_k) \enspace | \enspace (t_i, p_k) \in R,\enspace p_k \in P_m \} $ and similar for the algorithm's output\\$ \hat{R}_a := \{ (t_i, p_k) \enspace | \enspace (t_i, p_k) \in R_a,\enspace p_k \in P_m \} $ can be defined.
Again, identical mappings are calculated with the intersection $\hat{R}_m := \hat{R} \cap \hat{R}_a $.
Doing this for all persons $\forall p_k \in P_m$, we can calculate the average ($avg$) precision and recall for finding the relationships:
$$prec_R := avg(\frac{|\hat{R}_m|}{|\hat{R}_a|})  \qquad recall_R := avg(\frac{|\hat{R}_m|}{|\hat{R}|})$$

Third, our goal is to find out how well the algorithm detects ambiguity.
Similar to the person index comparison, the ground truth's list $A$ is compared with the algorithm's list $A_a$.
However this time, we individually consider every person that was correctly suggested in the group of ambiguous people.
Formally, since an element $(t_i, r, P_A) \in A$ contains a set of ambiguous people $P_A$, we define an auxiliary set $\hat{A} := \{ (t_i, r, p_A) \enspace | \enspace (t_i, r, P_A) \in A,\enspace p_A \in P_A \}$ to ease further comparison.
Again, an intersection can be calculated as $ \hat{A}_m := \hat{A} \cap \hat{A}_a $.
With this, the precision and recall for ambiguity detection can be stated as follows:
$$ prec_A := \frac{|\hat{A}_m|}{|\hat{A}_a|}  \qquad recall_A := \frac{|\hat{A}_m|}{|\hat{A}|} $$

For each precision and recall pair, we can calculate the harmonic mean which is commonly known as the F-score value:
$$ fscore := 2 * \frac{prec * recall}{prec + recall} $$

Our actual evaluation can be divided into two parts.
In the first part, we will examine how our method's features effect its detection performance.
By testing on various generated data, we discovered that only the first name gazetteer improves results in average.
In the second part, we will investigate eight experiments with specifically generated datasets.
They gave insights in our method's detection performance in more detail. 
Whenever we generate ground truth, we took for its input the 100 most common last names in USA\footnote{https://en.wikipedia.org/wiki/List\_of\_most\_common \_surnames\_in\_North\_America\#United\_States\_.28American.29 (Accessed 2020-04-05)} and collected 197 popular first names\footnote{https://en.wikipedia.org/wiki/List\_of\_most\_popular \_given\_names\#Americas (Accessed 2020-04-05)} (without middle names).

Because our approach has three usable features, we first examine how they have an effect on its detection performance.
The described features are (a) clearing adaptive data, (b) usage of probability measures and (c) lookup in a first name gazetteer list.
Since these features influence the detection of persons, we investigate how accurately our algorithm compiles the person index using the $f_P$ measurement.
Therefore, several tests with differently generated data are performed where no persons with middle names and no ambiguity are involved.
60 ground truth datasets of various sizes were generated at random.
In particular, we vary the number of persons $|P|$ (10 to 100), the amount of short texts $|T|$ (150 to 1500), the maximum number of persons mentioned in a text snippet (1 to 10) and the random seed.
Without any feature activated, the algorithm reaches in average an $f_P$ of $0.369\pm0.012$.
The usage of probability measures to filter improbable detections does not show any effect.
If for every new short text adaptive data is cleared, the value slightly reduces to $0.358\pm0.026$.
Once a first name gazetteer is used, the correction of name assignments increases $f_P$ in average to $0.462\pm0.024$.
Because of these insights, in further evaluation only the first name gazetteer feature is used.

Next, we will examine in more detail how our algorithm performs on eight generated datasets.
\autoref{tbl:eval} summarizes the evaluation results of the experiments.
Regarding ground truth, 
$|P|$ denotes the length of the person index list, 
$|T|$ shows the number of generated short texts,
\textit{Max} is the maximum number of persons mentioned in a text,
\textit{MN} counts how many persons have a middle name and
\textit{Amb} states the degree of ambiguity.
With an ambiguity degree of $n$, the dataset will have $n$ groups of $n$ people each share a last name while other $n$ groups of $n$ persons each share a first name.
As stated in the evaluation section, the precision and recall measures are calculated accordingly.
In the following, the eight experiments are examined.

In the first experiment, our algorithm correctly found the one person in ten different variations because the generation patterns always contain either first name or last name.
However, if the number of generated short texts is increased (experiment number two), more persons are incorrectly detected. 
This is mainly due to role names and department names that look like real names (e.g. ``Chief'', ``Admin'', etc).
Since so many falsely extracted persons share equal names, the algorithm assumes that they are all ambiguous and does not state any correct relation between short texts and persons in $R_a$.

In the third test, we increase the number of persons to 20, still, per short text only one person is mentioned.
Because different names are used, the algorithm has a better chance to find a correct name pair in the patterns.
This is also the reason why more correct relations are found, since there is less possibility for ambiguity.

For the fourth run, the maximum number of mentioned persons per text is increased to 10, thus now multiple persons can be mentioned in one text.
This situation seems to distract our algorithm which results in a lower $f_P$ and $f_R$ value.  
Regarding line 5, the performance declines slightly if persons with middle names are introduced.
In fact, only one individual with a middle name was identified correctly.
We assume that the OpenNLP model is not trained to identify persons with middle names.

In the sixth experiment, ambiguity is introduced:
two groups of two people each share a last name while other two groups of two persons each share a first name.
All eight ambiguous persons where discovered correctly in $P_a$ and their ambiguity were detected completely in $A_a$ (since $recall_A$ reaches $1.00$).
However, many other persons are incorrectly listed in the index which share first names and last names by accident. 
This results in a very low $prec_A$ precision.
If the degree of ambiguity is increased (line 7) and nearly every person is ambiguous (18 of 20), the recall declines a little.

In the last experiment, we increase the number of persons to 40 and short texts to 300.
Having more persons, the algorithm falsely assumes more ambiguity which declines $recall_A$ slightly.

\begin{table*}[t]
	\centering
	\caption{
		Performance of the baseline algorithm on generated ground truth data. 
		$|P|$ -- length of person index list,
		$|T|$ -- number of generated short texts,
		\textit{Max} -- maximum number of mentioned persons per text,
		\textit{MN} -- number of persons with middle names and
		\textit{Amb} -- degree of ambiguity.
		Precision (\textit{prec}), recall and f-score (\textit{f}) are calculated based on the relations $P$, $R$ and $A$.
	}
	\resizebox{\textwidth}{!}{\begin{tabular}{|c|r|r|r|r|r|| r|r|r|| r|r|r|| r|r|r|}
		\hline
		Nr. & $|P|$ & $|T|$ & Max & MN & Amb & $prec_P$ & $recall_P$ & $f_P$ & $prec_R$ & $recall_R$ & $f_R$ & $prec_A$ & $recall_A$ & $f_A$ \\
		\hline
		\hline
		1 & 1 & 10 & 0 & 0 & 0 & 1.00 & 1.00 & 1.00 & 1.00 & 1.00 & 1.00 & - & - & - \\
		\hline
		2 & 1 & 200 & 0 & 0 & 0 & 0.14 & 1.00 & 0.25 & 0.00 & 0.00 & - & - & - & - \\
		\hline
		3 & 20 & 200 & 0 & 0 & 0 & 0.63 & 0.85 & 0.72 & 0.82 & 0.72 & 0.77 & - & - & - \\
		\hline
		4 & 20 & 200 & 10 & 0 & 0 & 0.38 & 0.90 & 0.54 & 0.61 & 0.09 & 0.16 & - & - & - \\
		\hline
		5 & 20 & 200 & 10 & 4 & 0 & 0.31 & 0.80 & 0.45 & 0.63 & 0.09 & 0.16 & - & - & - \\
		\hline
		6 & 20 & 200 & 10 & 4 & 2 & 0.39 & 0.75 & 0.52 & 0.47 & 0.41 & 0.44 & 0.03 & 1.00 & 0.06 \\
		\hline
		7 & 20 & 200 & 10 & 4 & 3 & 0.45 & 0.85 & 0.59 & 0.59 & 0.54 & 0.56 & 0.05 & 0.95 & 0.09 \\
		\hline
		8 & 40 & 300 & 10 & 4 & 3 & 0.39 & 0.75 & 0.52 & 0.53 & 0.49 & 0.51 & 0.03 & 0.87 & 0.06 \\
		\hline
	\end{tabular}
	}
	\label{tbl:eval}
\end{table*}

Our experiments show that our baseline does not reach an $f_P$ value above $0.6$ in more realistic use cases.
Also $f_R$ reveals similar poor results.
Since many wrong persons with same names are discovered, $prec_A$ lists the worst results.
As improvements, we suggest to train detection models that are able to distinguish first name and last name.  
In addition, more context should be considered when relations between texts and persons are discovered.
For example, a letter of an abbreviated name can reduce ambiguity drastically.
We assume that if more persons are correctly discovered, the precision of ambiguity discovery will increase.

\section{\uppercase{Conclusion and Outlook}}
\label{sec:concl}

\noindent
In this paper, we asked how well an unsupervised algorithm is able to build a person index from a set of short texts.
We defined a person index as a structured table that distinctly catalogs individuals by their names.
After we gave a formal definition for this problem, we proposed a generator that is able to produce ground truth datasets for this challenge inspired by concrete data.
Additionally, a first baseline approach was described to examine first results.
In the evaluation, several measurements were defined to examine the performance of potential solutions.
With this, we analyzed our approach and suggested further potentials for improvement for future approaches.

For future work, we plan to examine performance in real use cases using data of our industrial scenarios.
In case of ambiguities, our goal is to efficiently integrate human experts which are able to contribute with their knowledge. 
The challenge itself can be made more difficult by generating names with different cases (i.e. lower case, upper case, mixed case, camel case, etc).
Regarding the domain, we aim to generalize the problem statement to other entity types which have multiple names or IDs in different forms.

\bibliographystyle{apalike}
{\small \bibliography{paper}}

\end{document}